\documentclass[wcp]{jmlr}

\usepackage[utf8]{inputenc} 
\usepackage[T1]{fontenc}    
\usepackage{hyperref}       
\usepackage{url}            
\usepackage{booktabs}       
\usepackage{amsfonts}       
\usepackage{nicefrac}       
\usepackage{microtype}      
\usepackage{natbib}

\usepackage{algorithm,algorithmic}
\usepackage{mathtools}
\usepackage{slashbox}
\usepackage{multirow}
\usepackage{multicol}

\title[Capsule Networks Need an Improved Routing Algorithm]{Capsule Networks Need an Improved Routing Algorithm}

  \author{\Name{Inyoung Paik} \Email{iypaik@deepbio.com}\\
  \Name{Taeyeong Kwak} \Email{tykwak@deepbio.com}\\
  \addr Deep Bio Inc., Seoul, Republic of Korea
  \AND
  \Name{Injung Kim} \Email{ijkim@handong.edu}\\
  \addr Handong Global University, Pohang, Republic of Korea
 }

\begin{document}
	
	\maketitle
	
	\begin{abstract}
		
		In capsule networks, the routing algorithm connects capsules in consecutive layers, enabling the upper-level capsules to learn higher-level concepts by combining the concepts of the lower-level capsules. Capsule networks are known to have a few advantages over conventional neural networks, including robustness to 3D viewpoint changes and generalization capability. However, some studies have reported negative experimental results. Nevertheless, the reason for this phenomenon has not been analyzed yet. We empirically analyzed the effect of five different routing algorithms. The experimental results show that the routing algorithms do not behave as expected and often produce results that are worse than simple baseline algorithms that assign the connection strengths uniformly or randomly. We also show that, in most cases, the routing algorithms do not change the classification result but polarize the link strengths, and the polarization can be extreme when they continue to repeat without stopping. In order to realize the true potential of the capsule network, it is essential to develop an improved routing algorithm.
		
	\end{abstract}

	\section{Introduction}
	
     \citep{capsule_original} and \citep{capsule_EM} introduced a novel deep learning architecture called \textit{capsule network}, which was inspired by the mechanism of the visual cortex in biological brains. The capsule network aims to achieve equivariance to variations, such as pose change, translation, and scaling. A capsule is composed of a group of neurons, each of which, hopefully, represents an attribute of the object. The orientation of a capsule denotes the pose of an object, while the length of the capsule represents the probability of the object's existence. To make this possible, the capsule network uses \textit{routing algorithms} to determine the link strength between each capsules. The connection links between the capsules in consecutive layers represent the part-whole relations among the objects represented by the capsules.  
	
	 \citep{capsule_EM} extends the concept of  \citep{capsule_original} to separate the activation of capsules from their poses. The agreement between capsules in consecutive layers contributes to strengthening the activation of the higher capsules; the connection between two capsules becomes fortified as they are activated together, as in the case of the Hebbian rule. The concept of the expectation-maximization procedure is employed to estimate the connection strength between capsules.
	
	After the first proposition, many new routing algorithms \citep{capsule_spectral, capsule_BiGRU, capsule_ms, capsule_group, capsule_pro, capsule_attn, capsule_optim, capsule_seg} and applications \citep{capsule_lungcancer, capsule_unsupervised, capsule_3d} have been proposed, and results that surpass those of existing CNNs have been reported. However, some studies have reported negative results on reinforcement learning \citep{complex_DQN} and image classification \citep{complex, complex2}. It is still necessary to investigate issues such as what the routing algorithm really does and what it can and cannot do.
	
	
	Although many experiments have been conducted on routing algorithms, there have been few experiments that directly show the effect and behavior of the routing algorithm. For example, most of the previous studies limited the number of routing iterations to a small number, such as 3. In this study, we performed a detailed analysis of the behavior of the two original routing algorithms \citep{capsule_EM, capsule_original} as well as three other algorithms that were proposed recently \citep{capsule_optim, capsule_group, capsule_attn}. For convenience, in this paper, we will refer to each network as follows: \textit{CapsNet} \citep{capsule_original}, \textit{EMCaps} \citep{capsule_EM}, \textit{OptimCaps} \citep{capsule_optim}, \textit{GroupCaps} \citep{capsule_group}, and \textit{AttnCaps} \citep{capsule_attn}. 
	
	\section{Routing Algorithms}
	In a capsule network, neurons are grouped into units called capsules, and the routing algorithm determines the strength of the links between them by iterations of update steps. Although the detailed formulae of the routing algorithms differ, routing algorithms roughly work as follows:\bigskip
	
	1. Determine the initial output value of capsules($\vec{y_j}^{(0)}$) from the values of the lower-level capsules.\\
	2. Calculate the 'agreement'($b_{ij}$) between the input($\vec{x_i}$) and output($\vec{y_j}$) capsules.\\
	3. Reinforce the link strength between 'agreed' capsules ($c_{ij}^{(t)} \rightarrow c_{ij}^{(t+1)}$).\\
	4. Recalculate the value of the output capsules using the updated link strength ($y_{j}^{(t)} \rightarrow y_{j}^{(t+1)}$).\\
	5. Repeat steps 2-4. \bigskip

	Usually, the initial value of an output capsule is calculated from uniformly assigned link strengths (except \textit{AttnCaps} \citep{capsule_attn}, where the initial output is calculated from separate input). For each routing cycle, the upper capsules compute their activation based on the current link strengths. Then, the routing algorithm measures the agreement between capsules and updates the link strength based on the degree of agreement. Agreement between capsules can be calculated by the inner product \citep{capsule_original, capsule_optim, capsule_attn, capsule_lungcancer}, Gaussian probability \citep{capsule_EM, capsule_3d, capsule_spectral}, or distance measure \citep{capsule_group}. Once the link strengths have been updated, the cycle starts again. In practice, routing is repeated only a fixed number of times (usually three) rather than until convergence.

	\begin{algorithm}
		\caption{The routing-by-agreement algorithm by  \citep{capsule_original} (\textit{CapsNet})}
		\begin{algorithmic}[1]
			\scriptsize
			\STATE $\vec{x}_{i}$ : inputs \\
			\STATE $W_{ij}$ : weights \\
			\STATE $\vec{u}_{ij} = W_{ij}\vec{x}_{i}$\\
			\STATE $b_{ij} \leftarrow 0$ \\
			\STATE for $r$ iterations do :\\
			\STATE $\quad c_{ij} \leftarrow \frac{exp(b_{ij})}{\sum_{k}exp({b_{ik}})}$\\
			\STATE $\quad \vec{s}_{j} \leftarrow \sum_{k}c_{kj}\vec{u}_{kj}$ \\
			\STATE $\quad \vec{y}_{j} \leftarrow \frac{\Vert \vec{s}_{j} \Vert \vec{s}_{j}}{1+\Vert \vec{s}_{j} \Vert ^{2} } $ \\
			\STATE $\quad b_{ij} \leftarrow b_{ij} + \vec{u}_{ij} \cdot \vec{y}_{j}$ \\
			\STATE $return \enspace \vec{y}_{j} $ \\
		\end{algorithmic}
	\end{algorithm}

	\begin{algorithm}
		\caption{Expectation-maximization routing algorithm by  \citep{capsule_EM} (\textit{EMCaps})}
		\begin{algorithmic}[1]
			\scriptsize
			\STATE $ X_{i}, a_{i}$ : inputs \\
			\STATE $W_{ij}, \beta_{1}, \beta_{2}$ : weights \\
			\STATE $\lambda, \epsilon$ : hyperparameter \\
			\STATE $U_{ij} = W_{ij}X_{i}$\\
			\STATE ${u}_{ij}^{h} :=$ h-th component of $ U_{ij}$\\
			\STATE $b_{ij} \leftarrow 1/(number \enspace of\enspace upper\enspace capsules)$ \\
			\STATE for $r$ iterations do :\\
			\STATE $\quad c_{ij} \leftarrow b_{ij}a_{i}$\\
			\STATE $\quad y_{j}^{h} \leftarrow \frac{\sum_{k}c_{kj}u_{kj}^{h}}{\sum_{k}c_{kj}} $\\
			\STATE $\quad (\sigma_{j}^{h})^{2} \leftarrow \frac{\sum_{k}c_{kj}(u_{kj}^{h} - y_{j}^{h})^{2}}{\sum_{k}c_{kj}} + \epsilon$\\
			\STATE $\quad cost_{j}^{h} \leftarrow \sum_{h}(\beta_{1} + log(\sigma_{j}^{h}))\sum_{k}c_{kj}$\\
			\STATE $\quad a_{j}' \leftarrow logistic(\lambda(\beta_{2}-\sum_{h}cost_{j}^{h}))$ \\
			\STATE $\quad s_{ij} \leftarrow (\prod_{h} 2\pi(\sigma_{j}^{h})^{2})^{-1/2} exp(-\sum_{h}\frac{(u_{ij}^{h} - y_{j}^{h})^{2}}{2 (\sigma_{j}^{h})^{2}})$\\
			\STATE $\quad b_{ij} \leftarrow \frac{a_{j}' s_{ij}}{\sum_{k} a_{k}' s_{ik}}$\\
			\STATE $return \enspace Y_j, a'_{j} $ \\
		\end{algorithmic}
	\end{algorithm}

	\begin{algorithm}
		\caption{Routing-by-agreement algorithm by  \citep{capsule_optim} (\textit{OptimCaps})}
		\begin{algorithmic}[1]
			\scriptsize
			\STATE $\vec{x}_{i}$ : inputs \\
			\STATE $W_{ij}$ : weights \\
			\STATE $\lambda$ : hyperparameter \\
			\STATE $\vec{u}_{ij} = W_{ij}\vec{x}_{i}/\Vert W_{ij} \Vert_{\textit{F}}$ * Frobenius norm\\
			\STATE $b_{ij} \leftarrow 0$ \\
			\STATE for $r$ iterations do :\\
			\STATE $\quad c_{ij} \leftarrow \frac{exp(b_{ij})}{\sum_{k}exp({b_{ik}})}$\\
			\STATE $\quad \vec{s}_{j} \leftarrow \sum_{k}c_{kj}\vec{u}_{kj}$ \\
			\STATE $\quad \vec{v}_{j} \leftarrow \frac{\vec{s}_{j}}{\Vert \vec{s}_{j} \Vert } $ \\
			\STATE $\quad b_{ij} \leftarrow \lambda \vec{u}_{ij} \cdot \vec{v}_{j}$ \\
			\STATE $w_{j} = \frac{\Vert \vec{s}_{j} \Vert}{1+max_{k}\Vert \vec{s}_{k} \Vert}$\\
			\STATE $\vec{y}_{j} = w_{j}\vec{s}_{j}$\\
			\STATE $return \enspace \vec{y}_{j} $ \\
		\end{algorithmic}
	\end{algorithm}

	\begin{algorithm}
		\caption{Group equivariant routing algorithm by  \citep{capsule_group} (\textit{GroupCaps})}
		\begin{algorithmic}[1]
			\scriptsize
			\STATE $\vec{x}_{i}, a_{i}$ : inputs \\
			\STATE $W_{ij}, \alpha, \beta$ : weights \\
			\STATE $M, \delta$ : weighted average operator, distance measure \footnotemark  \\
			\STATE $\vec{u}_{ij} = W_{ij}\vec{x}_{i}$\\
			\STATE $\vec{y}_{j} \leftarrow M(\vec{u}_{kj}, a_{k})$ \\
			\STATE for $r$ iterations do :\\
			\STATE $\quad c_{ij} \leftarrow sigmoid(-\alpha \delta( \vec{y_{j}}, \vec{u}_{ij}) + \beta)$\\
			\STATE $\quad \vec{y}_{j} \leftarrow M( \vec{u}_{ij}, c_{ij})a_{i}$ \\
			\STATE $a_{j}'= sigmoid(-\alpha \frac{1}{(\# \enspace input \enspace capsules)}\sum_{k} \delta ( \vec{y_{j}}, \vec{u}_{kj} ) + \beta)$\\
			\STATE $return \enspace \vec{y}_{j},a_{j}' $ \\
		\end{algorithmic}
	\end{algorithm}
	\footnotetext{Please refer to  \citep{capsule_group} for an exact definition.}

	\begin{algorithm}
		\caption{Routing algorithm with attention by  \citep{capsule_attn} (\textit{AttnCaps})}
		\begin{algorithmic}[1]
			\scriptsize
			\STATE $\vec{x}_{i}, \vec{h}_{j}$ : inputs \\
			\STATE $W_{j}, M_{j}$ : weights \\
			\STATE $\vec{u}_{ij} = W_{j}\vec{x}_{i}$\\
			\STATE $\vec{y}_{j} \leftarrow M_{j}\vec{h}_{j}$\\
			\STATE $b_{ij} \leftarrow 0$ \\
			\STATE for $r$ iterations do :\\
			\STATE $\quad c_{ij} \leftarrow \frac{exp(b_{ij})}{\sum_{k}exp({b_{ik}})}$\\
			\STATE $\quad \vec{y}_{j} \leftarrow \vec{y}_{j} + \sum_{k}c_{kj}\vec{u}_{kj}$ \\
			\STATE $\quad b_{ij} \leftarrow b_{ij} + \vec{u}_{ij} \cdot \vec{y}_{j}$ \\
			\STATE $return \enspace \vec{y}_{j}, c_{ij} $ \\
		\end{algorithmic}
	\end{algorithm}

	\section{Questions and Empirical Verification}
	
	It is challenging to observe what really happens inside a neural network and to examine the positive and negative effects of new methods. We carefully planned controlled experiments to answer the following questions about the routing algorithms. These questions are listed from the most challenging one to the least challenging ones.\bigskip
	
	
	(Q1) Is the routing algorithm \textit{generally} better than the simple weighted sum operation?\\
	There have been a few \textit{general} improvements in deep learning. For example, using batch normalization \citep{BN} or residual connection \citep{resnet} in image classification  \textit{generally} leads to improvement, almost regardless of the choice of model and dataset. Does the routing algorithm of the capsule network provide such general improvement?
	
	Many previous studies have reported results using the capsule network that surpass those of baseline CNNs. However, since the architecture and computational complexity of the capsule network differ significantly from those of ordinary CNNs, this is still an open question. Since our study focuses on the routing algorithm rather than the overall effect of the capsule network, we directly compared between \textit{using} and \textit{not using} the routing algorithm without changing the architecture of the capsule networks to minimize the interference of other factors. Specifically, we implemented two alternative routing algorithms named \textit{uniform routing} and \textit{random routing} for each of the five routing algorithms. The uniform routing algorithm assigns all link strengths uniformly and never updates. In this case, the capsule network behaves similarly to ordinary neural networks, as the inference is mainly performed by the weighted sum operations. In random routing, link strength($c_{ij}$) is sampled from a random distribution ($U(0.8, 1.2)$) at every iteration, making the inference stochastic.
	
	The experimental results are presented in Table 1. All routing algorithms exhibited the best performance when they were repeated 2$\sim$3 times. Increasing the number of routing iterations did not improve the performance, but slightly degraded the performance.	Unexpectedly, the best performances of most routing algorithms were not superior to those of the uniform or random routing algorithms. The only exception was GroupCaps, whose best performance was slightly superior to those of the uniform and random routing algorithms.\bigskip

	(Q2) Does the routing algorithm help the capsule network achieve equivariance or invariance to 3D transforms such as 3D pose change, and thereby improves robustness to 3D transform? 
	
	The capsule network was motivated by a 3D pose change matrix to obtain equivariance to 3D viewpoint changes \citep{capsule_EM}. In order to minimize the influence of other factors such as a complex background, we experimented on the smallNORB \citep{norb} dataset, which contains gray-scale images of a few objects taken at many different angles, with a carefully designed pure shape recognition task not disturbed by context or color \citep{capsule_EM}.
	
	The results are presented in Table 2. As in the previous experiment, all routing algorithms exhibited superior performance when they were repeated 3 times rather than 10 times. None of the routing algorithms outperformed the uniform and random routing algorithms.\bigskip
	
	(Q3) Does the routing algorithm help the capsule network achieve equivariance or invariance to 2D transform such as translation and rotation, thereby improving robustness to 2D transform?
	
	To evaluate robustness to 2D rotation, we experimented on the rotated MNIST \citep{MNIST} dataset; In this experiment, we trained the model by rotating the training data randomly within the [-30, 30] degree range. We then tested with randomly rotated images over a wider range of [-180, 180] degrees. We also ran a similar experiment for translation by placing the $28 \times 28$ digit images at random coordinates on $36 \times 36$ white images. The results are exhibited in Table 3 and Table 4. In both cases, the routing algorithms did not exhibit any improvement compared to the uniform routing algorithm. \bigskip
	
	(Q4) Is the capsule network heavily reliant on the routing algorithm, or is it capable of adapting to a variety of routing algorithms?
	
	In general, neural networks with sufficient capacity can learn tolerance to noise or variation. Moreover, adding noise in training often improves the generalization capability. From this perspective, it is possible that the capsule network is not very sensitive to the choice of routing algorithm unless the routing algorithm plays a key role, since it can adapt to the connection links even when they are not optimally assigned.
	
	To verify this, we conducted an experiment that uses the routing algorithms normally during training and uses the uniform or random routing algorithms during evaluation. The results are presented in the 5th and 7th columns of Table 2. Surprisingly, replacing the routing algorithms with the uniform or random routing algorithms did not severely degrade, but rather improved the performance of the capsule networks that were trained with the routing algorithms. One possible reason is that the five routing algorithms tend to overly polarize the link strengths as described in Section 5, while the uniform and random routing algorithms make the distribution of the link strength less sharp, allowing the network represent uncertainty. \bigskip
	
	Additionally, we conducted an experiment to know how frequently the routing algorithm changes the classification result. We trained \textit{CapsNet} \citep{capsule_original} on MNIST dataset \citep{MNIST} for few epochs, and compared two experimental classifiers : one selects the class whose output capsule was assigned the largest initial activation value $u_j = \sum_i u_{ij}$ which are computed before the routing of the output layer, and the other selects the class whose capsule was assigned with the largest link strength $c_j = \sum_i c_{ij}$ which is directly computed by the routing algorithm. Then, we compared the accuracy of the two classifiers. The results are presented in Figure 1. As learning progresses, the capsule with the largest initial activation value is also assigned with the largest link strength in most cases. This result suggests that the classification results are mainly determined by the initial activation values of the output capsules and rarely changed by the routing algorithm.
	

	
	\begin{table}[H]
	\footnotesize
		\begin{center}
			\begin{tabular}{l|l|l|l|l|l|l}
				\multirow{2}*{Algorithm} & \multicolumn{6}{|c}{Routing iteration}\\ \cline{2-7}
				& Uniform & Random & 2 iterations & 3 iterations & 5 iterations & 10 iterations\\ \hline
				\textit{OptimCaps} & 93.39(0.11) & 93.31(0.12) & 93.13(0.23) & 93.16(0.21) & 92.60(0.05) & 92.32(0.31)\\ \hline
				\textit{GroupCaps} & 91.36(0.84) & 91.27(0.61) & 91.19(1.16) & 91.51(0.79) & 90.90(0.95) & 90.29(0.68) \\ \hline
				\textit{CapsNet} & 90.86(0.29) & 91.09(0.13) & 88.93(0.11) & 88.41(0.16) & 87.40(0.13) & 86.49(0.14) \\ \hline
				\textit{EMCaps} & 93.03(0.01) & 93.17(0.16) & 92.76(0.18) & 92.56(0.17) & 92.27(0.14) & 92.35(0.22) \\ \hline
				\textit{AttnCaps} & 93.20(0.11) & 93.08(0.14) & 83.16(1.31) & 70.27(5.36) & 57.46(15.98) & 42.57(13.88) \\ 
			\end{tabular}
		\end{center}
		\caption{The accuracy of the five routing algorithms for CIFAR10 varying with the number of routing iterations, compared to the uniform and random algorithms. This is the result of the experiment to verify Q1. In the \textit{Uniform} option, the link strength between capsules is uniformly assigned and never updated. In the \textit{Random} option, the link strengths between capsules were sampled from $U(0.8, 1.2)$ each time. We found no evidence that any of the five algorithms generally improves the performance of the deep learning model. See section 4 for the experimental details.}
	\end{table}

	
	\begin{table}[ht]
	\footnotesize
		\begin{center}
			\begin{tabular}{l|l|l|l|l|l|l}
				\multirow{3}*{Algorithm} & \multicolumn{6}{|c}{Routing iteration}\\ \cline{2-7}
				& \multirow{2}*{Uniform} & \multirow{2}*{Random} & \multirow{2}*{3 iteration} & 3 iters + & \multirow{2}*{10 iterations} &10 iters +\\
				& & & & Uniform & & Random\\\hline
				\textit{OptimCaps} & 91.21(1.13) & 91.75(0.65) & 90.90(1.07) & 90.94(0.68) & 89.31(0.50) & 90.96(1.18)\\ \hline
				\textit{GroupCaps} & 90.05(0.56) & 91.64(0.83) & 90.14(1.17) & 89.96(0.07) & 89.19(0.44) & 89.67(0.36) \\ \hline
				\textit{CapsNet} & 91.93(1.05) & 90.93(1.05) & 91.11(0.40) & 91.39(0.96) & 91.67(0.33) & 91.16(0.93) \\ \hline
				\textit{EMCaps} & 90.73(0.43) & 91.17(0.42) & 91.17(0.43) & 91.61(0.96) & 90.54(1.01) & 91.23(0.53)\\ \hline
				\textit{AttnCaps} & 90.73(0.45) & 90.37(1.34) & 84.01(2.79) & 84.10(1.14) & 58.35(16.2) & 58.23(31.2) \\ 
			\end{tabular}
		\end{center}
		\caption{The accuracy of the routing algorithms for SmallNORB varying with the number of routing iterations. This is the result of the experiment to verify Q2 and Q4. The 5th and 7th columns display the results of the experiments in which the networks were trained using the corresponding routing algorithms, and then, evaluated using the uniform or random routing algorithms. The capsule network worked fine even if we used different routing algorithms for training and evaluation. These results suggest that the model can learn to produce consistent results no matter how the link strengths are assigned, similarly to as if it adapts to random perturbations like \citep{dropout, shakeshake}.}
	\end{table}

	\begin{table}[ht]
	\footnotesize
		\begin{center}
			\begin{tabular}{l|l|l|l|l|l|l}
				\multirow{3}*{Algorithm} & \multicolumn{6}{|c}{Routing iteration}\\ \cline{2-7}
				& \multicolumn{2}{|c|}{Uniform} &\multicolumn{2}{|c|}{3 iterations} & \multicolumn{2}{|c}{10 iterations}\\ \cline{2-7}
				& [-30, 30] & [-180, 180] & [-30, 30] & [-180, 180] & [-30, 30] & [-180, 180]\\ \hline
				\textit{OptimCaps} & 99.57(0.02) & 61.64(0.47) & 99.51(0.07) & 59.77(0.67) & 99.53(0.02) & 59.78(0.59) \\ \hline
				\textit{GroupCaps} & 99.47(0.04) & 62.06(0.66) & 99.44(0.04) & 57.76(1.65) & 99.43(0.07) & 60.32(0.48) \\ \hline
				\textit{CapsNet} & 99.60(0.01) & 61.68(0.65) & 99.51(0.04) & 59.07(0.57) & 99.42(0.05) & 59.09(1.01) \\ \hline
				\textit{EMCaps} & 99.56(0.04) & 60.43(0.34) & 95.21(2.12) & 52.32(1.67) & 88.16(1.60) & 45.47(1.63) \\ \hline
				\textit{AttnCaps} & 99.54(0.02) & 60.42(0.22) & 99.21(0.25) & 57.93(1.67) & 89.86(7.98) & 45.66(8.79) \\ 
			\end{tabular}
		\end{center}
		\caption{The accuracy of the routing algorithms for rotational MNIST dataset. This is the result of the experiment to verify Q3. All models were trained with images randomly rotated in the range of [-30, 30] degrees, and then, evaluated with images randomly rotated in the range of [-30, 30] degrees and [-180, 180] degrees. We found no evidence that the routing algorithms improve the generalization capability to 2d rotation.}
	\end{table}

	\begin{table}[ht]
	\footnotesize
		\begin{center}
			\begin{tabular}{l|l|l|l|l|l|l}
				\multirow{3}*{Algorithm} & \multicolumn{6}{|c}{Routing iteration}\\ \cline{2-7}
				& \multicolumn{2}{|c|}{Uniform} &\multicolumn{2}{|c|}{3 iterations} & \multicolumn{2}{|c}{10 iterations}\\ \cline{2-7}
				& Center & Random & Center & Random & Center & Random\\ \hline
				\textit{OptimCaps} & 99.67(0.02) & 97.46(0.24) & 99.59(0.04) & 95.09(0.84) & 99.52(0.10) & 96.19(0.53) \\ \hline
				\textit{GroupCaps} & 99.50(0.04) & 95.73(1.63) & 99.59(0.06) & 95.03(1.77) & 99.56(0.05) & 95.13(0.83) \\ \hline
				\textit{CapsNet} & 99.69(0.05) & 99.02(0.32) & 99.54(0.12) & 96.36(0.58) & 99.60(0.04) & 97.32(0.61) \\ \hline
				\textit{EMCaps} & 99.59(0.26) & 89.38(0.46) & 99.47(0.14) & 88.13(1.93) & 99.37(0.21) & 88.29(2.10) \\ \hline
				\textit{AttnCaps} & 99.58(0.05) & 95.44(0.85) & 99.39(0.06) & 93.91(1.70) & 98.55(0.61) & 79.81(7.77)\\ 
			\end{tabular}
		\end{center}
		\caption{The accuracy of the routing algorithms for translated MNIST dataset. This is the result of the experiment to verify Q3. In training, each $28 \times 28$ digit image was placed at the center of a $36 \times 36$ white image. In evaluation, it was put at the center or a random coordinate of the $36 \times 36$ white image. We found no evidence that the routing algorithms improve generalization capability to 2d translation.}
	\end{table}


	\begin{figure}[ht]
		\vskip 0.2in
		\begin{center}
			\centerline{\includegraphics[width=12cm]{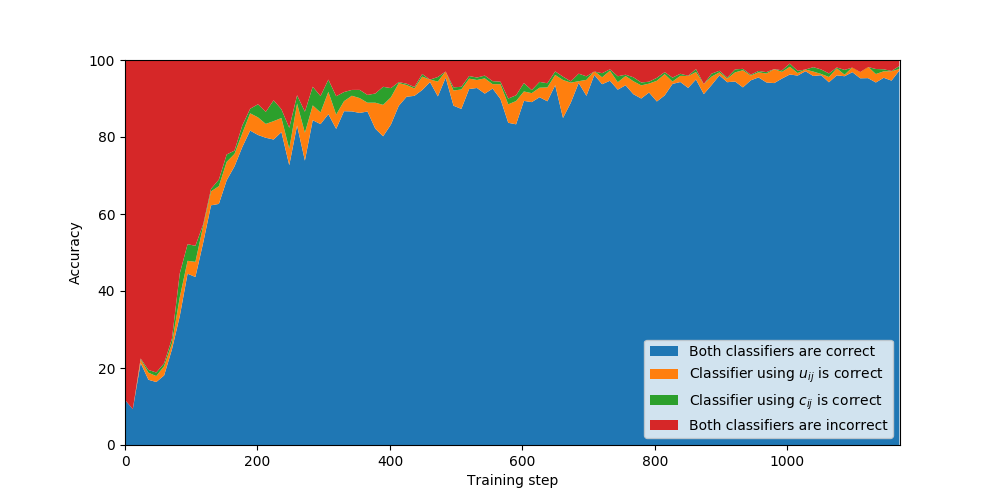}}
			\caption{
			 The accuracy of two experimental classifiers: one classifies using the initial activation values $u_j = \sum_i u_{ij}$  of the output capsules and the other using the link strengths $c_j=\sum_i c_{ij}$. This is the result of the experiment to know how frequently the routing algorithm changes the classification result. The blue region represents the ratio that both classifiers are correct. The orange and green regions represent the ratios that only one of them is correct. The red region represents the ratio that both classifiers are incorrect. At the end of the experiment, the portion of the four regions were 95.75\%, 1.21\%, 0.54\%, and 2.50\%, respectively. The capsule with the largest initial activation value is also assigned with the largest link strength in most cases, which suggests that the classification results are mainly determined by the initial activation values of the output capsules and rarely changed by the routing algorithm.}

			\label{fig:connections2}
		\end{center}
		\vskip -0.2in
	\end{figure}

	\section{Experimental details}
	
	\subsection{Experimental Settings}
	
	We used a ResNet-34 model  \citep{resnet} without a classification layer as a feature extractor, and added a capsule layer. Global average pooling was applied for  \citep{capsule_original, capsule_optim, capsule_attn}, but not for  \citep{capsule_EM, capsule_group}, which requires spatial information. The length of each capsule was fixed to 16. We set the batch size to 128 in most experiments, except in the multi-layer experiment of \textit{GroupCaps} \citep{capsule_group}, in which we used a batch size of 32 because of the excessive demand for GPU memory. We used the stochastic gradient descent optimizer with a momentum of 0.9. In all experiments, we trained for 50 epochs. The learning rate was initially set to 0.1 and divided by 10 at the 16th and 32nd epochs. We did not use any early stopping criteria. However, when the training failed completely (for example, training accuracy dropped to nearly 10\%), we restarted the experiment. For the CIFAR10 \citep{cifar} dataset, we applied random cropping with a padding size of 4 and random horizontal flip. We resized SmallNORB \citep{norb} images to $48 \times 48$ and randomly cropped to sample $32 \times 32$ regions. For the MNIST \citep{MNIST} dataset, we added padding to make $32 \times 32$ images. In our experiments, we set $\lambda=1$ for \textit{OptimCaps} and $\lambda=0.01$ for \textit{EMCaps}. We empirically searched for the value of $\lambda$ for \textit{OptimCaps} \citep{capsule_optim} and \textit{EMCaps} \citep{capsule_EM} from $10^{-4}$ to $10^4$.
	    
	We ran all experiments three times and present the average accuracy in Tables 1-4. The numbers in the parentheses denote the standard deviations. Since we used the same architecture for three different datasets, there was overfitting for smallNORB and MNIST. However, we did not attempt to minimize overfitting, because our experiments only aimed to make relative comparisons.

	\subsection{Minor Observations}
	
	\begin{itemize}
	
        \item We observed that for \textit{OptimCaps}, the accuracy was improved as we decreased $\lambda$, which leads the behavior of \textit{OptimCaps} closer to the uniform routing. The performance of \textit{OptimCaps} also converged to that of uniform routing as $\lambda$ approaches zero. We also observed similar tendency for \textit{EMCaps} \citep{capsule_EM}.
		
		\item We found that adding a large $\epsilon$ to variance($(\sigma^h_j)^2$) in \textit{EMCaps}  \citep{capsule_EM} stabilizes learning and prevents divergence. However, no matter how large an $\epsilon$ we added, the performances of the algorithm approached that of the uniform routing. We used $\epsilon=0.01$ for all experiments mentioned above.
		
		\item We mainly used a ResNet model composed of 34 layers in our experiments. We presume capsules are more appropriate to represent high-level concepts than low-level concepts such as lines and edges, because they are more informative and their part-whole relations are less ambiguous than those of low-level concepts. Nevertheless, we also attempted to use a shallow feature extractor composed of only two layers, but did not obtain improved results.
		
		\item We also experimented with multiple (2$\sim$5) capsule layers after the ResNet feature extractor, but did not obtain improved results. In addition, to verify whether the routing algorithm causes the capsule network to learn capsule-wise feature representation, i.e., whether the neurons in a capsule collectively represent the pose and existence of an entity or not, we attempted to disassemble the neurons in the capsules and reorganized them into capsules arbitrarily but fixed way. However, we did not observe any negative impact on the learning of the capsule network.
		
		\item We also attempted other techniques such as 1) adding reconstruction loss as \textit{CapsNet} \citep{capsule_original}, 2) pretraining with a different dataset, and 3) blocking the gradient from $c_{ij}$, in order to hide the routing algorithm from the neural network. However, we did not obtain meaningful improvement.
		
		
	\end{itemize}

	\section{Polarization Problem}
	
	In the experiments, the routing algorithms tended to overly polarize the link strengths, increasing the strength of a small number of connections while suppressing all other connections, as shown in Figure 2. In particular, when they were repeated multiple times, the routing algorithm almost always produced a simple route in which each input capsule passed its value to only one output capsule and all other routes were suppressed. Here we analyzed the reason for the polarization issue for  \citep{capsule_original}. The behavior of the other routing algorithms can be analyzed similarly.
	
	Let a connection $b_{ij}$ and its corresponding softmax output $c_{ij}$ be stronger than $b_{ik}$ and $c_{ik}$ for all $k \neq j$, respectively. This draws $\vec{s}_{j}$ closer to $\vec{u}_{ij}$, and as a result, the routing algorithm increases $b_{ij}$ and $c_{ij}$ more rapidly than the other routes. Since the algorithm does not have any mechanisms to control the feedback loop, $b_{ij}$ increases infinitely if the routing algorithm is not stopped, drawing $c_{ij}$ to 1 and all other routes $c_{ik}$ to 0 for all $k \neq j$.
	
	\begin{figure}[ht]
		\vskip 0.2in
		\begin{center}
			\centerline{\includegraphics[width=12cm]{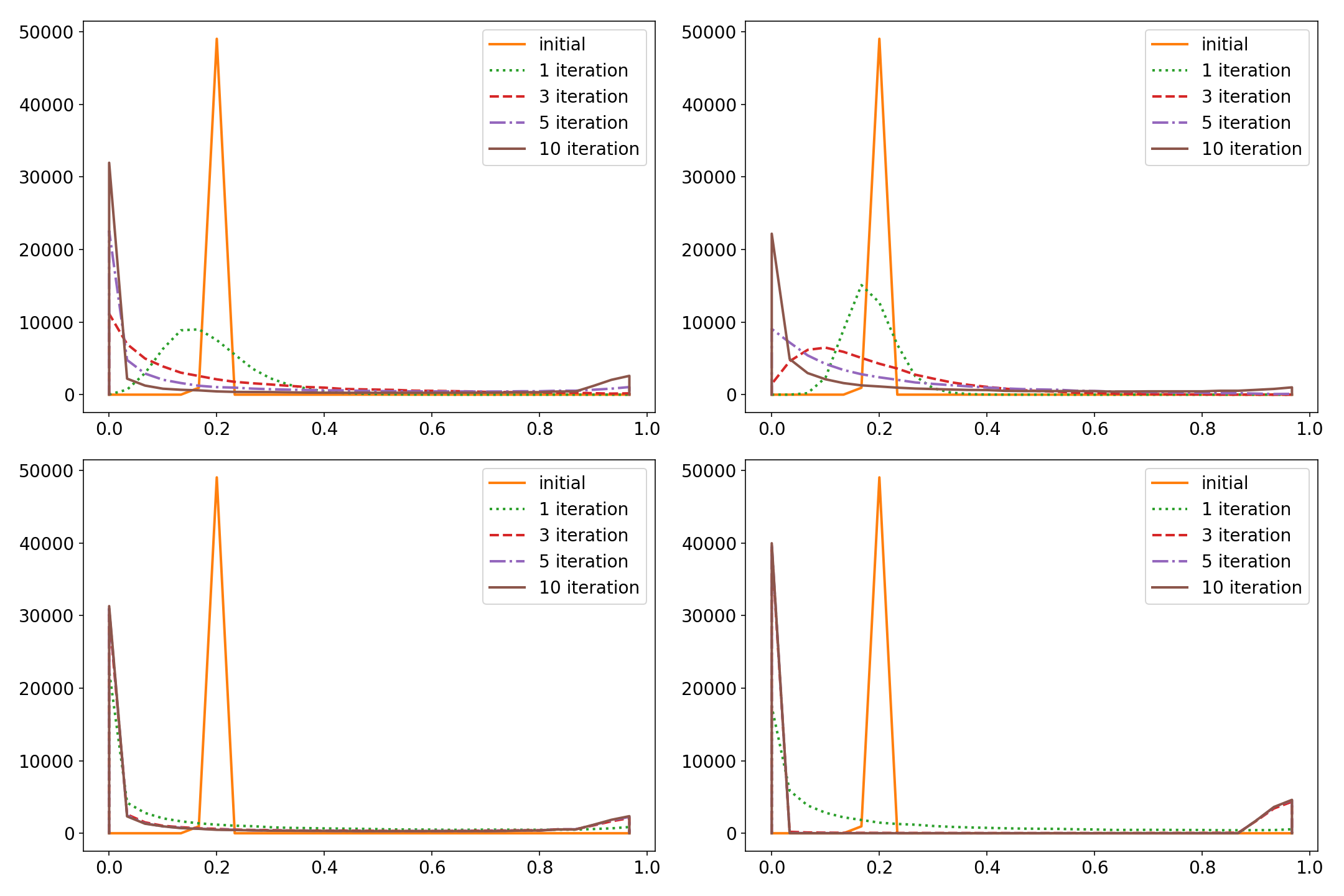}}
			\caption{Histogram of link strength ($c_{ij}$) of five output capsules produced by the routing algorithms varying with the number of routing iterations. \textit{CapsNet} \citep{capsule_original} (left top), \textit{OptimCaps} \citep{capsule_optim} (right top), \textit{EMCaps} \citep{capsule_EM} (left bottom), and \textit{AttnCaps} \citep{capsule_attn} (right bottom) \label{fig:connections} As the routing iteration increases, the link strengths were polarized more severely. When the routing was repeated 10 times, the most of the links fell to zero and only a small number of links increased to 1.}
		\end{center}
		\vskip -0.2in
	\end{figure}
	
	Not surprisingly, this is the only realistic steady state of the algorithm, i.e., the only plausible destination if it converges. It could be even worse if the algorithm does not lead to convergence, considering the purpose of the routing algorithm. However, we did not observe divergence or oscillation in the routing algorithms. Let $c_{ij}^{(t)}$ be the value of $c_{ij}$ at the $t$-th routing iteration and $a_{ij} = \vec{u}_{ij} \cdot \vec{y}_{j}$. Then,\bigskip
	
	\begin{equation}
		\begin{aligned}
			c_{ij}^{(t+1)} &= \frac{exp(b_{ij}^{(t+1)})}{\sum_{k}exp(b_{ik}^{(t+1)})} = \frac{exp(b_{ij}^{(t)})exp(a_{ij}^{(t)})}{\sum_{k}exp(b_{ik}^{(t)})exp(a_{ik}^{(t)})} \\
			&= \frac{exp(b_{ij}^{(t)})exp(a_{ij}^{(t)})/\sum_{r}exp(b_{ir}^{(t)})}{\sum_{k}exp(b_{ik}^{(t)})exp(a_{ik}^{(t)})/\sum_{r}exp(b_{ir}^{(t)})} = \frac{c_{ij}^{(t)}exp(a_{ij}^{(t)})}{\sum_{k}c_{ik}^{(t)}exp(a_{ik}^{(t)})} 
		\end{aligned}
	\end{equation}
	
	If the algorithm reaches a steady state, $c_{ij}^{(t+1)} = c_{ij}^{(t)}$, Eq. (2) holds.
	
	\begin{equation}
		\{ exp(a_{ij}^{(t)}) = \sum_{k}c_{ik}^{(t)}exp(a_{ik}^{(t)}) \}\enspace or \enspace \{ c_{ij}^{(t)}=0 \} \enspace \forall (i,j)
	\end{equation}
	
	Let ${r_{i}}$ be the number of output capsules with $c_{ij}^{(t)} > 0$. If $r_{i} = 1$, there exists a simple solution $c_{ij} = 1$ and $c_{ik} = 0$ for all $k \neq j$, which requires that $b_{ij}$ diverges to infinity or that all the other routes collapse to 0. This is the what we always observed. If $r_{i} \geq 2$, Eq. (2) requires a condition $a_{ij_{1}} = a_{ij_{2}} = ... = a_{ij_{r_{i}}}$. However, it is not only undesirable state of routing algorithm but also highly unlikely that the routing algorithm naturally produces such a singular solution. \textit{OptimCaps} \citep{capsule_optim} and \textit{AttnCaps} \citep{capsule_attn} share the same problem, since they compute $c_{ij}$ similarly.\bigskip
	
	In the case of \textit{EMCaps} \citep{capsule_EM}, the reason of the polarization is less obvious because the computation of the link strength is more complicated. However, similar to the routing-by-agreement algorithm, we observed that repeating the \textit{EMCaps} \citep{capsule_EM} algorithm multiple times also led to a simple solution, allowing a single non-zero outing route from each capsule. Let there be some $i_{0}$, $u_{i_{0}j} \simeq \mu_j$ and suppose $\sigma_j \simeq 0$. Then, $s_{i_{0}j}$, $b_{i_{0}j}$, and $c_{i_{0}j}$ are larger than the others, bringing $\mu_{j}$ closer to $u_{i_{0}j}$ and reducing $\sigma_{j}$. In the end, the distribution approaches a delta function centered at $u_{i_{0}j}$, and the likelihood diverges to infinity. 
	
	This is similar to the known singularity problem in the Gaussian mixture model, but this problem seems to appear more serious than usual because the number of data points (number of capsules in the pre-layer) per distribution (number of capsules in the post-layer) is extremely low - only nine times the number of distributions, for a 3x3 convolutional capsule layer, as in \textit{EMCaps} \citep{capsule_EM}.\bigskip

	Finding a simple solution or even a singular solution is a natural consequence of the routing algorithms, as they were designed to reinforce the connection between the capsules that agree with each other. However, such solutions are undesirable for guiding the capsule network to learn part-whole relations, because they make the network overconfident and deny uncertainty.

	\section{Conclusion}
	
	We empirically analyzed five recently developed routing algorithms for the capsule network. The experimental results suggest that the routing algorithms have yet to realize the potential of the capsule network. We also showed that the routing algorithms overly polarize the link strengths, and this issue can be extreme when they continue to repeat without stopping. In order to achieve the design goals of the capsule network, we believe it is essential to continue searching for an improved routing algorithm that has a mathematical foundation for ensuring convergence to a desirable state allowing for uncertainty in part-whole relations rather than polarizing the link strengths.
	


	\bibliographystyle{spmpsci}      
	

\end{document}